%% file: root.tex
\documentclass[letterpaper, 10 pt, journal, twoside]{IEEEtran}

\IEEEoverridecommandlockouts                              

\usepackage{amsmath,amsfonts}
\usepackage{algorithm}
\usepackage{algpseudocode}
\usepackage{array}
\usepackage[caption=false,font=normalsize,labelfont=sf,textfont=sf]{subfig}
\usepackage{textcomp}
\usepackage{stfloats}
\usepackage{url}
\usepackage{verbatim}
\usepackage{graphicx}
\usepackage{cite} 
\usepackage{xcolor}

\title{
RESPRECT: Speeding-up Multi-fingered Grasping with Residual Reinforcement Learning
}

\author{Federico~Ceola,
        Lorenzo~Rosasco
        and~Lorenzo~Natale
\thanks{Manuscript received: September 6, 2023; Revised: December 13, 2023; Accepted: January 19, 2024.}
\thanks{This paper was recommended for publication by Editor Aleksandra Faust upon evaluation of the Associate Editor and Reviewers' comments.
This work was supported by the PNRR MUR project PE0000013-FAIR. L. R. acknowledges the financial support of the European Research Council (grant SLING 819789), the European Commission (ELIAS 101120237), the US Air Force Office of Scientific Research (FA9550-18-1-7009 and FA8655-22-1-7034), the Ministry of Education, University and Research (grant ML4IP R205T7J2KP) and the Center for Brains, Minds and Machines (CBMM), funded by NSF STC award CCF-1231216. This work represents only the view of the authors. The European Commission and the other organizations are not responsible for any use that may be made of the information it contains.}
\thanks{Federico Ceola and Lorenzo Natale are with Humanoid Sensing and Perception (HSP), Istituto Italiano di Tecnologia (IIT), Genoa, Italy (email: {\tt\footnotesize name.surname@iit.it}).}
\thanks{Federico Ceola and Lorenzo Rosasco are with Laboratory for Computational and Statistical Learning (LCSL), with Machine Learning Genoa Center (MaLGa) and with Dipartimento di Informatica, Bioingegneria, Robotica e Ingegneria dei Sistemi (DIBRIS), University of Genoa, Genoa, Italy.}
\thanks{Lorenzo Rosasco is also with Istituto Italiano di Tecnologia (IIT) and with Center for Brains, Minds and Machines (CBMM), Massachusetts Institute of Technology (MIT), Cambridge, MA (email: {\tt\footnotesize lrosasco@mit.edu}).}
}

\begin{document}

\maketitle


\begin{abstract}
Deep Reinforcement Learning (DRL) has proven effective in learning control policies using robotic grippers, but much less practical for solving the problem of grasping with dexterous hands -- especially on real robotic platforms -- due to the high dimensionality of the problem.

In this work, we focus on the multi-fingered grasping task with the anthropomorphic hand of the iCub humanoid. We propose the \textit{RESidual learning with PREtrained CriTics} (RESPRECT) method that, starting from a policy pre-trained on a large set of objects, can learn a residual policy to grasp a novel object in a fraction ($\sim 5 \times$ faster) of the timesteps required to train a policy from scratch, without requiring any task demonstration. To our knowledge, this is the first Residual Reinforcement Learning (RRL) approach that learns a residual policy on top of another policy pre-trained with DRL. We exploit some components of the pre-trained policy during residual learning that further speed-up the training. We benchmark our results in the iCub simulated environment, and we show that RESPRECT can be effectively used to learn a multi-fingered grasping policy on the real iCub robot.

The code to reproduce the experiments is released together with the paper with an open source license\footnote{\label{repo_fn}\url{https://github.com/hsp-iit/rl-icub-dexterous-manipulation}}.

\end{abstract}



\section{Introduction}
\label{sec:introduction}
\input{sections/introduction.tex}


\section{Related Work}
\label{sec:relwork}
\input{sections/relwork.tex}


\section{Methodology}
\label{sec:methods}
\input{sections/methods.tex}


\section{Experimental Setup}
\label{sec:exp_setup}
\input{sections/exp_setup.tex}


\section{Results}
\label{sec:results}
\input{sections/results.tex}


\section{Limitations}
\label{sec:limitations}
\input{sections/limitations.tex}


\section{Conclusion}
\label{sec:conclusions}
\input{sections/conclusions}


\appendices

\section{}
\label{appendix:ft}
\input{sections/appendix_ft}

\section{}
\label{appendix:reptile}
\input{sections/appendix_reptile}

\section{}
\label{appendix:gpayn}
\input{sections/appendix_gpayn}

\section{}
\label{appendix:residual}
\input{sections/appendix_residual}


\bibliographystyle{unsrt}
\bibliography{bibliography}

\end{document}

%% file: sections/introduction.tex
\IEEEPARstart{L}{earning} dexterous manipulation tasks is an open challenge in robotics~\cite{andrychowicz2020learning}. These tasks require controlling tens of degrees of freedom (DoFs), to deal with possibly imprecise perception of the environment, and to manage hand-object interactions. Grasping is the key task to enable the execution of other dexterous manipulation tasks such as object re-orientation~\cite{andrychowicz2020learning, chen2022system}.

The latest model-free DRL approaches, such as SAC~\cite{haarnoja2018soft} or PPO~\cite{schulman2017proximal}, are feasible options to learn problems with high-dimensionality. However, their deployment on real robotic platforms is difficult due to the huge amount of timesteps required to explore the environment at the beginning of the training. A major trend in the literature to overcome this problem is to train a policy in simulation and transfer it on the real robot~\cite{qin2023dexpoint}. However, the success of these approaches depends on the sim-to-real gap between the simulated and the real environments. With our work, we aim at overcoming this limitation by proposing a method for fast learning of the multi-fingered grasping task.

The contribution of this paper is a method that we call RESPRECT. The core of this algorithm is an RRL method that leverages on a pre-trained base policy to learn a residual additive policy on a novel object. In contrast to existing residual learning methods, which usually rely on classical closed-loop controllers as base policies, we show that it is possible to learn a residual policy also if a hand-tuned closed-loop base controller is unavailable, as for the considered multi-fingered grasping task.

As a further contribution, we propose to exploit pre-trained components to speed-up the residual training. We initialize the SAC \emph{Critics} components in the residual policy using the weights of the pre-trained policy. In the experimental section we show that this leads to a significant reduction of the training time, making it suitable for learning policies directly in the real world.

We benchmark our results against conventional fine-tuning and Meta Reinforcement Learning (MetaRL) approaches in a simulated environment with the 9-DoFs hand of the iCub humanoid~\cite{icub}, surpassing all the considered baselines in most of the experiments. Furthermore, the experimental results show that RESPRECT achieves the same success rate as the state-of-the-art method for multi-fingered grasping G-PAYN~\cite{ceola2023gpayn}, while requiring only a fraction of the training timesteps ($1M$ compared to $5M$). Notably, differently from G-PAYN~\cite{ceola2023gpayn}, RESPRECT accomplishes this without the need for grasping demonstrations to initialize the training process. While these requirements prevent G-PAYN~\cite{ceola2023gpayn} from learning grasping policies on the real robot, we deploy RESPRECT on the real iCub, showing that it can learn a grasping policy for two new objects in $\sim2.5$ hours and $\sim30$ minutes.

To the best of our knowledge, this is the first DRL algorithm that has been successfully used to solve the problem of grasping with an articulated hand with several DoFs directly on the real robot from visual, tactile and proprioceptive data, and without the need for task demonstrations.

%% file: sections/relwork.tex
\textbf{Multi-fingered Grasping} The literature mainly focuses on grasp poses generation~\cite{vezzani2017sq} or detection of finger-object contact points~\cite{10115005, li2022efficientgrasp}, whereas the deployment of efficient control strategies to perform the grasp has received less attention. Recent work proposes to solve the problem with DRL-based approaches. The work in~\cite{liang2021multifingered_rl} starts from a grasp pose given by an external algorithm, and relies on synergies to reduce the number of DoFs to control the fingers of a Shadow hand. The policy is trained on a multi-modal input comprising tactile information, joint angles and torques, which are not always available in other robotic hands. Moreover, it does not take into account information about the object (e.g., visual feedback of the environment or object pose) during grasp execution to allow grasp recovery if the initial grasp pose is unfeasible. G-PAYN~\cite{ceola2023gpayn}, instead, considers as proprioceptive information the cartesian pose of the end-effector and finger joint positions, relying on visual feedback from the head-mounted camera of the robot. However, training is performed in simulation with long training sessions relying on huge amounts of grasping demonstrations, and therefore it cannot be performed on the real robot. DexPoint~\cite{qin2023dexpoint}, instead, learns to grasp and open a door from pointclouds, showing that sim-to-real transfer can be obtained without fine-tuning. While being an interesting research direction, this transfer is strongly affected by sensors quality. This is particularly evident with objects in proximity to depth sensors. We overcome these limitations providing a method that can be deployed on a real robot for fast learning of multi-fingered grasping from visual, tactile, and proprioceptive data. Notably, differently from the state-of-the-art G-PAYN~\cite{ceola2023gpayn}, RESPRECT does not require any task demonstration, while being trained much faster.

\textbf{Fast DRL} State-of-the-art DRL algorithms, e.g. SAC~\cite{haarnoja2018soft} or PPO~\cite{schulman2017proximal}, are sample-inefficient and require huge amounts of training episodes to be optimized, hindering their application on tasks that require to be trained on real robots. Some methods adapt standard DRL algorithms leveraging off-line task demonstrations to overcome this limitation.~\cite{vecerik2017leveraging, nair2018overcoming} adapt DDPG~\cite{lillicrap2015continuous} to exploit task demonstrations: DDPGfD~\cite{vecerik2017leveraging} leverages on a prioritized replay mechanism to sample transitions between demonstrations and agent data, while~\cite{nair2018overcoming} adds a Behavior Cloning (BC) loss component to the one of DDPG~\cite{lillicrap2015continuous} to mimic demonstrations. 
DAPG~\cite{Rajeswaran-RSS-18}, instead, fine-tunes a DRL policy pre-trained on demonstrations with imitation learning. Off-line DRL approaches as AWAC~\cite{nair2020awac}, instead, pre-train a policy with DRL on demonstrations and then adapt it on-line on the robot. Results in~\cite{ceola2023gpayn} show that these approaches are not suited for multi-fingered grasping from visual, tactile, and proprioceptive data. A different approach to adapt a DRL policy is fine-tuning.~\cite{zhao2022effectiveness} compares fine-tuning to several MetaRL approaches. While being promising for fast adaptation of DRL tasks, MetaRL algorithms are often difficult to use in practice due to their complexity. For example, PEARL~\cite{rakelly2019efficient} requires learning an additional network for task context inference, which is used to condition the trained policy. In some cases, MetaRL methods can be employed only with on-policy DRL algorithms~\cite{finn2017model}. We benchmark our method against fine-tuning and MetaRL baselines.

\textbf{RRL} RRL methods aim at speeding-up policy learning. They are designed to predict a residual action, that is added to the output of an existing controller. The first RRL approach was introduced in~\cite{silver2018residual} to improve imperfect controllers in simulated manipulation tasks, such as object pushing or pick-and-place.~\cite{johannink2019residual, schoettler2020deep} propose RRL methods for insertion tasks on real robots, either from observable and measurable states, or from raw pixels.~\cite{ranjbar2021residual} proposes to modify also the signal to a base feedback controller to avoid the feedback distribution shift caused by the residual policy, which the base controller tries to resist. These approaches share a common limitation, in that they rely  on manually designed conventional controllers, which are difficult to design for a multi-fingered grasping task.~\cite{davchev2022residual} overcomes this limitation learning a residual policy to solve insertion tasks on top of Dynamic Movement Primitive (DMP) base policies extracted through BC. However, learning DMPs from visual, tactile and proprioceptive data is impractical. \cite{alakuijala2021residual} proposes an RRL method to improve a policy trained with BC by superimposing a residual policy trained with DRL on seven simulated manipulation tasks (the same tasks used for BC training). This removes the dependency on hand-engineered base controllers, but requires task demonstrations which are difficult to obtain on the real robot. This challenge becomes even more pronounced in the context of this work, where we aim at learning multi-fingered grasping policies on objects unseen during base policy training. Furthermore, evidence from previous work~\cite{ceola2023gpayn} shows that training policies on the task at hand with BC is impractical. Residual learning has also found application on different robotic tasks, such as object throwing~\cite{zeng2020tossingbot}, where the residual throwing velocity is regressed and superimposed on the velocity predicted by an ideal physics controller, or to learn navigation strategies~\cite{9197386}, where a classical controller serves as the base policy.

\noindent We overcome the limitations of the state-of-the-art, which either depend on classical base controllers (unavailable for the considered multi-fingered grasping task), or rely on a policy trained with BC on the same task. The goal of the proposed algorithm is to allow the robot to quickly learn to grasp unseen objects, starting from a DRL policy that is pre-trained on a set of generic objects. To achieve this, we use visual features obtained from a backbone pre-trained on a dataset including egocentric images from everyday tasks, without making any assumption on the target object to be grasped. We leverage the pre-trained DRL policy even further: the residual \emph{Critics} component is initialized using the pre-trained weights of the base policy, significantly speeding-up the training. These enhancements are inherently unattainable using the base policies employed in existing literature.

%% file: sections/methods.tex
\subsection{Grasping Pipeline}
\label{sec:grasp_pipeline}

We rely on the grasping pipeline introduced in~\cite{ceola2023gpayn}, and consider the right hand of the iCub humanoid robot. This is actuated by 9 motors and is equipped with tactile sensors on the fingertips. The pipeline includes two main phases. In the initial phase, the end-effector is moved in a pose spaced $5cm$ from a grasp pose generated by an object agnostic algorithm. This is either an algorithm based on superquadrics~\cite{vezzani2017sq}, hereinafter referred to as \textit{Superquadrics}, or VGN~\cite{breyer2020volumetric}. Subsequently, we employ a DRL policy, trained with the proposed RESPRECT, to approach and lift the object. The DRL policy controls the $6$ DoFs of the end-effector's pose and the $9$ finger joints to approach the object and finally lift it.

We consider five elements as the state of the Markov Decision Process (MDP) underlying the DRL policy at hand. We use the visual Flare~\cite{shang2021reinforcement} features extracted with the \textit{ViT-Large} model from the Masked Autoencoder presented in~\cite{radosavovic2023real} (MAE). This model was trained on several real-world visual datasets, comprising Ego4D~\cite{grauman2022ego4d} that well represents the type of visual feedback acquired by the robot head-mounted camera. We also consider a binary tactile value for each fingertip, the cartesian pose of the end-effector, finger joint positions and an estimate of the initial pose of the object to grasp (the latter computed as the center of the superquadric, or the center of the segmented pointcloud, when considering VGN for grasp pose synthesis). The DRL policy outputs a $15$-dimensional vector that represents offsets for moving the end-effector and finger joints. The policy is trained with a reward function that considers: the pose of the hand with respect to the estimated initial position of the object, the number of fingers in contact with the object (detected in the case of tips-object meshes contact in simulation, or when tactile sensors on the real robot are triggered), the position of the object when lifted, and the terminal condition of the episode. An episode is regarded as successfully terminated if the object is grasped and uplifted by $10cm$. Conversely, an episode is deemed a failure if the object is moved too far from the initial position, the inverse kinematics (IK) solver cannot find a solution for the specified configuration, or the number of timesteps exceeds the designated maximum threshold. For further details, we refer the reader to the description in~\cite{ceola2023gpayn}.

\begin{figure*}
    \vspace{1mm}
    \centering
    \includegraphics[width=\linewidth]{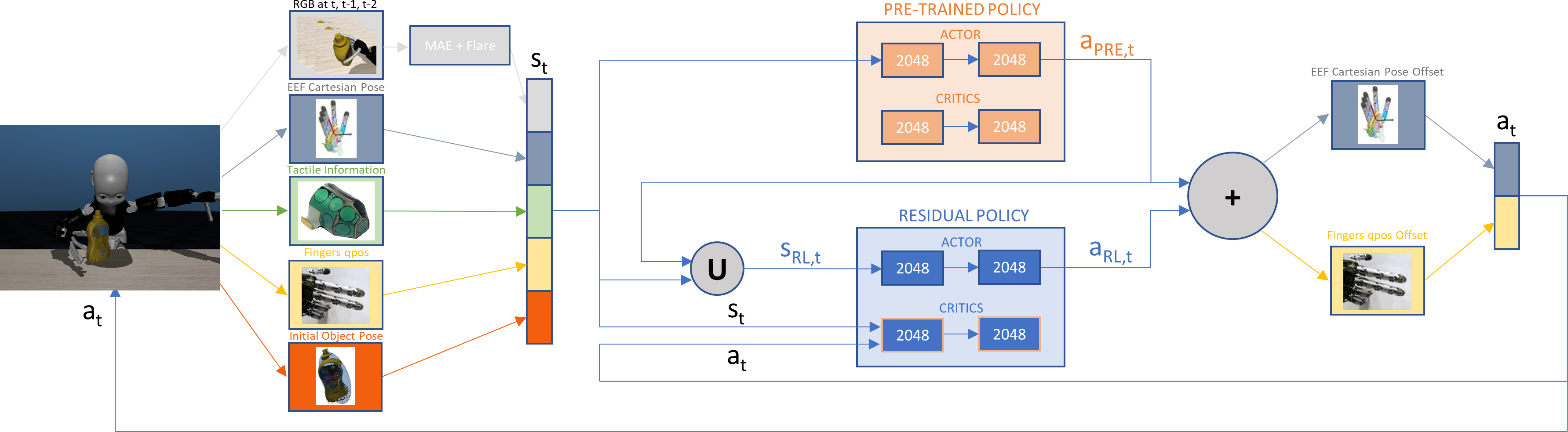}
    \caption{RESPRECT overview. We compute state $s_t$ from RGB images at timesteps $t$, $t-1$ and $t-2$ (processed through the MAE in~\cite{radosavovic2023real} and combined with Flare~\cite{shang2021reinforcement}), end-effector cartesian pose, tactile information and finger joint poses. We compute action $a_t$ (composed of cartesian offsets for the end-effector and finger joint offsets) combining the outputs $a_{PRE, t}$ of the pre-trained policy and $a_{RL, t}$ of the residual policy. Note that $a_{RL, t}$ is the output of the residual policy \textit{Actor}, given the concatenation of $s_t$ and $a_{PRE, t}$ into $s_{RL, t}$. We train only the two $2048$-dimensional fully connected layers in the residual \textit{Actor} and \textit{Critics}. For the latter, we start from the \textit{Critics} weigths of the pre-trained policy (orange outline). For the sake of clarity, we do not report the input of the \textit{Critics} in the pre-trained policy, and the output of both the \textit{Critics} being the same as the ones in SAC~\cite{haarnoja2018soft}.}
    \label{fig:pipeline}
\end{figure*}

\subsection{Residual Policy Training}
\label{subsec:residual_policy_training}
To train the grasping policy we propose a novel RRL method. This leverages on two components: a policy that is pre-trained with G-PAYN~\cite{ceola2023gpayn} in simulation, and a residual, object-specific, policy trained with a modified version of SAC~\cite{haarnoja2018soft}. The former is trained for two million timesteps using the MuJoCo models of the \textit{Scanned Objects Dataset} (MSO)~\cite{downs2022scannedobjects, zakka2022scannedobjectsmujoco}. In the simulated experiments, the latter is trained on seven YCB-Video~\cite{xiang2018posecnn} objects, while in the real world experiment it is trained on the \textit{006\_mustard\_bottle} and on the \textit{021\_bleach\_cleanser} (also taken from YCB-Video). Given the state of the system $s_t$ at time $t$, our approach outputs an action $a_t$ for the robot. This is the sum of two components: the action $a_{PRE, t}$ predicted by the pre-trained policy, and the action $a_{RL, t}$ predicted by the residual policy. While training the residual policy, the weights of the pre-trained policy remain fixed, and this is used only to predict $a_{PRE, t}$ which is needed for the optimization of the residual policy. As shown in Fig.~\ref{fig:pipeline}, we modify the standard SAC~\cite{haarnoja2018soft} inputs for the \textit{Actor} and \textit{Critics} components in the residual policy. Specifically, for the \textit{Actor}, we compute the state $s_{RL, t}$ as the concatenation between the state $s_t$ and the action produced by the pre-trained policy $a_{PRE, t}$. This allows the policy to compute the residual action $a_{RL, t}$ conditioned not only to the current state of the system, but also to the action produced by the pre-trained component. The \textit{Critics} are fed with the state $s_t$ and the action $a_t$, instead of $a_{RL, t}$ and $s_{RL, t}$ as in the standard SAC~\cite{haarnoja2018soft} algorithm. This allows training the residual \textit{Critics} starting with the weights of the pre-trained counterparts and to speed-up the initial stage of the training as demonstrated by the experiments.

%% file: sections/exp_setup.tex
We validate our approach in the MuJoCo~\cite{todorov2012mujoco} simulated environment customized for the iCub robot in~\cite{ceola2023gpayn}. We train all the policies by adapting the SAC~\cite{haarnoja2018soft} implementation in the Stable-Baselines3~\cite{stable-baselines3} library. For RESPRECT, we consider training hyperparameters as the ones in~\cite{ceola2023gpayn}, but we increase the number of gradient steps to $10$ and we set the initial entropy coefficient for SAC~\cite{haarnoja2018soft} to $0.01$. We provide a complete overview of the training hyperparameters in Tab.~\ref{tab:sac_params}.

\begin{table}[t]
    \vspace{2mm}
    \centering
    \begin{tabular}{c|c}
        Parameter                               & Value \\
        \hline
        Optimizer                               & Adam \\
        Learning Rate                           & $3 \cdot 10^{-4}$\\
        Discount ($\gamma$)                     & $0.99$\\
        Replay Buffer size                      & $10^6$\\
        Number of Hidden Layers (all networks)  & $2$\\
        Number of Hidden Units per Layer        & $1024$\\
        Number of Samples per Minibatch         & $256$\\
        Entropy Target                          & $-15$\\
        Non-linearity                           & ReLU\\
        Target Smoothing Coefficient ($\tau$)   & $0.005$\\
        Target Update Interval                  & $1$\\
        Gradient Steps                          & $10$\\                
        Training Frequency                      & $10$ Timesteps\\
        Total Environment Timesteps             & $1 \cdot 10^6$\\
        Entropy Coefficient                     & $0.01$\\
    \end{tabular}
    \caption{\textbf{RESPRECT} Training Hyperparameters.}
    \label{tab:sac_params}
\end{table}

\begin{figure}
    \centering
    \includegraphics[width=1\linewidth]{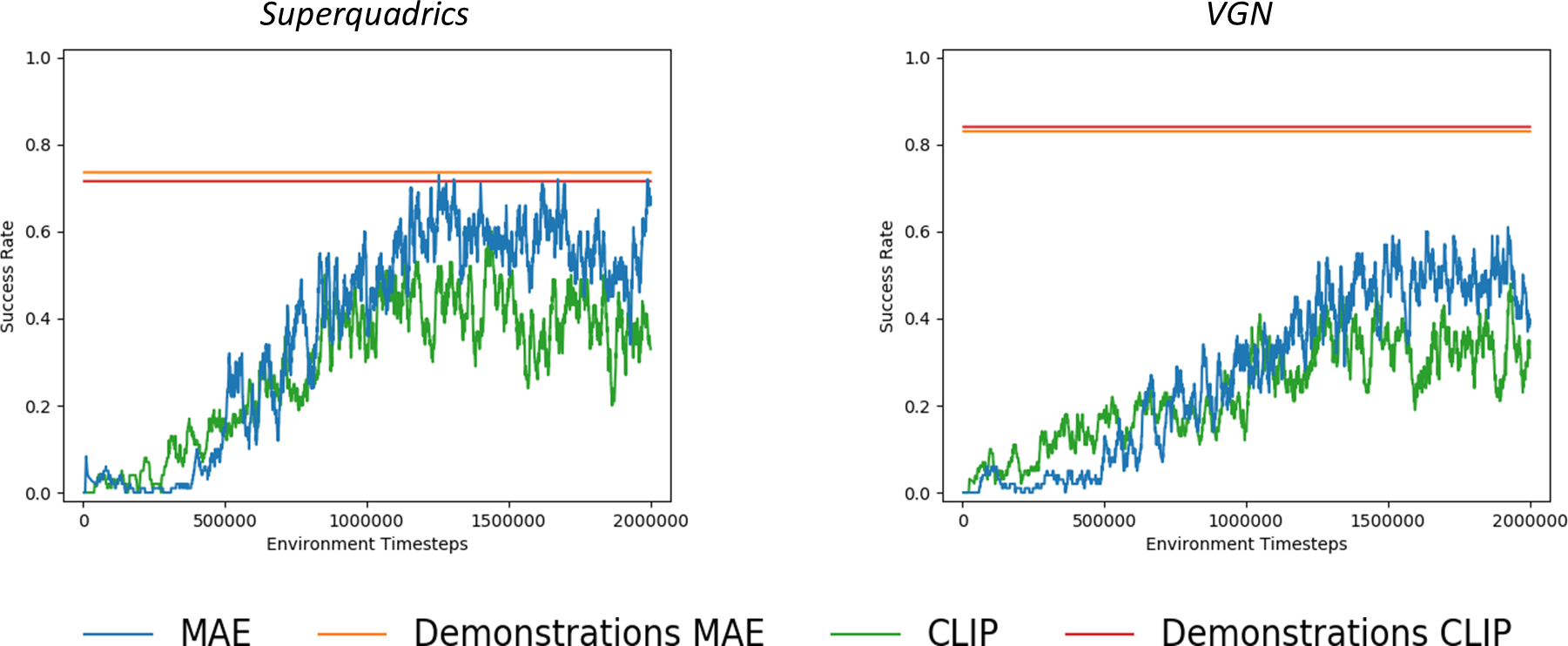}
    \caption{We compare the success rate obtained with different visual backbones (MAE and CLIP) when learning with G-PAYN~\cite{ceola2023gpayn} on the MSO dataset for $2M$ timesteps. We report in separate plots the cases in which we use \textit{Superquadrics} and VGN for the initial grasp pose synthesis. We also report the average success rate of the \textit{Demonstrations} used to initialize the G-PAYN replay buffer. Note that they slightly differ for CLIP and MAE due to the random initialization of each episode to collect demonstrations.}
    \label{fig:clip_mae_mso}
\end{figure}

Both for RESPRECT and the baselines (see Sec.~\ref{subsec:baselines}), we improve the visual feature extractor with respect to~\cite{ceola2023gpayn} by replacing the pre-trained \textit{ViT-B/32} CLIP~\cite{radford2021learning} model with the pre-trained \textit{ViT-Large} model of the MAE in~\cite{radosavovic2023real} and we increase the number of hidden units in the $2$ layers of the SAC~\cite{haarnoja2018soft} \textit{Actor} and \textit{Critics} to $2048$. The reason for this change is that the \textit{ViT-Large} model of the MAE led to higher success rate of the pre-trained policy. In Fig.~\ref{fig:clip_mae_mso} we compare G-PAYN trained on MSO with the two different feature extractors.

\subsection{Real Robot Setup}
We deploy our method on the real iCub\footnote{We run the module for training the grasping policy on a machine equipped with an Intel(R) Core(TM) i7-9750H CPU @ 2.60GHz, and a single NVIDIA RTX 2080 Ti.}~\cite{icub} humanoid. The robot is equipped with an \textit{Intel(R) RealSense D405} on a headset for the acquisition of RGB images and depth information (the latter is not used by the DRL algorithm but only during the approach phase to compute the initial grasp pose). We rely on the YARP~\cite{Mettayarp} middleware for the implementation and the communication between the different modules. For policy training, we adapt some components of the simulated training pipeline:
\begin{itemize}
    \item We use as input RGB image to the MAE the central crop of size $320\times240$ of the image acquired by the camera of the robot to match the real and simulated visual fields of view.
    \item We adapt some components of the reward function and some of the terminal conditions, since the precise position of the target object is difficult to obtain on the real robot. Specifically, to reward the position of the object along the axis perpendicular to the table ($r_{object\_height}$ in~\cite{ceola2023gpayn}), we consider the z-component of the position of the end-effector of the robot, once it has touched the object with at least two fingers. While this is sufficient to evaluate whether an episode ends positively (i.e. the object has been grasped), it does not allow to determine failure cases when the object falls off the table or moves away from its initial condition. We overcome this problem by manually sending a notification to the learning module.
    \item We assume that a finger is in contact with the object when the tactile sensors mounted on the fingertip is triggered. If a fingertip is in contact with other parts of the environment (e.g. the table or other fingers in possibly dangerous configurations) we consider this as a possibly unsafe state for the robot, and we manually terminate the execution of the grasping episode.
    \item We move the end-effector of the robot both to initialize the grasping (i.e. in a pose spaced $5cm$ from the one estimated by the \textit{Superquadrics}) and during policy execution via a cartesian controller that performs IK and trajectory computation~\cite{pattacini2010experimental}. We set the fixation point of the gaze of the robot to the center of the object, which is randomly placed on the table in a graspable configuration at the beginning of the grasping episode, as it is estimated by the \textit{Superquadrics}.
\end{itemize}

We refer the reader to the code released together with the paper for the implementation details of RESPRECT on the real iCub.

%% file: sections/results.tex
\begin{figure*}
    \vspace{1mm}
    \centering
    \includegraphics[width=1\linewidth]{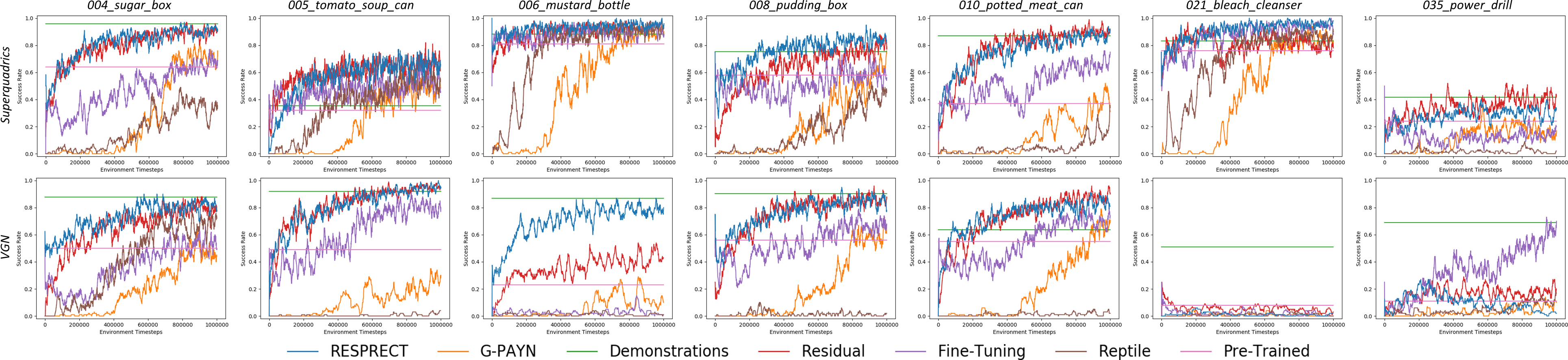}
    \caption{\textbf{Results}. We compare the success rate achieved by \textbf{RESPRECT} to the baselines for $1M$ environment timesteps. We benchmark the performance over seven YCB-Video objects (on different columns) starting from grasp poses generated either by \textit{Superquadrics} or \textit{VGN} (on different rows).}
    \label{fig:results}
\end{figure*}

To evaluate the effectiveness of our approach, we benchmark our results in the scenarios proposed in \cite{ceola2023gpayn}, using seven objects from the YCB-Video dataset chosen to represent various grasp types (the \textit{004\_sugar\_box}, the \textit{005\_tomato\_soup\_can}, the \textit{006\_mustard\_bottle}, the \textit{008\_pudding\_box}, the \textit{010\_potted\_meat\_can}, the \textit{021\_bleach\_cleanser}, and the \textit{035\_power\_drill}). As in \cite{ceola2023gpayn}, for each object, we employ two different methods for the computation of the initial grasp pose: \textit{Superquadrics} and \textit{VGN}. We then evaluate our approach training a policy to grasp a \textit{006\_mustard\_bottle} and a \textit{021\_bleach\_cleanser} on the real iCub robot.

\subsection{Baselines}
\label{subsec:baselines}

We benchmark \textbf{RESPRECT} (blue line in Fig.~\ref{fig:results}), comparing the success rate for increasing environment timesteps against the following:

\begin{itemize}
    \item \textbf{G-PAYN} (orange line in Fig.~\ref{fig:results}): this is the approach proposed in~\cite{ceola2023gpayn}. For a fair comparison, we train from scratch \textbf{G-PAYN} on the considered objects using the pre-trained MAE~\cite{radosavovic2023real} as feature extractor, and we increase the dimension of the two fully connected layers in the SAC \textit{Actor} and \textit{Critics} networks to $2048$.
    \item \textbf{Demonstrations} (green line in Fig.~\ref{fig:results}): this is the pipeline proposed in~\cite{ceola2023gpayn} that is used to collect the demonstrations for the training of  \textbf{G-PAYN}. We report the success rate of the demonstrations collected to fill the initial replay buffer in the \textbf{G-PAYN} experiment.
    \item \textbf{Residual} (red line in Fig.~\ref{fig:results}): this is a similar approach to \textbf{RESPRECT}, but we do not initialize the SAC \textit{Critics} of the residual policy with the weights of the SAC instance pre-trained on MSO. We provide an overview of this approach in App.~\ref{appendix:residual}. Note that \textbf{Residual} is a contribution itself over pre-existing methods that rely on classical base controllers, which, to the best of our knowledge are unavailable for the considered multi-fingered grasping task.
    \item \textbf{Fine-Tuning} (violet line in Fig.~\ref{fig:results}): we start from the policy pre-trained with \textbf{G-PAYN} on the MSO dataset and we fine-tune the fully connected layers in the \textit{Actor} and \textit{Critics} on the considered YCB-Video object. Differently from \textbf{RESPRECT} and \textbf{Residual}, we perform one gradient step at each training iteration. In App.~\ref{appendix:ft}, we compare the obtained success rate to the one achieved when performing $10$ gradient steps.
    \item \textbf{Reptile} (brown line in Fig.~\ref{fig:results}): this is an adaptation for the SAC~\cite{haarnoja2018soft} algorithm of the meta learning approach proposed in~\cite{nichol2018first}. We modify the original \textit{Reptile} algorithm as in~\cite{zhao2022effectiveness}, filling the initial replay buffers for the pre-training on MSO as in  \textbf{G-PAYN}. We chose this method because, according to the results shown in~\cite{zhao2022effectiveness}, it is the most competitive among the considered MetaRL baselines on the robotic benchmark RLBench~\cite{james2020rlbench}. As for \textbf{Fine-Tuning}, we perform one gradient step at each training iteration (see App.~\ref{appendix:reptile}). 
    \item \textbf{Pre-Trained} (pink line in Fig.~\ref{fig:results}): we compute the success rate achieved by the policy pre-trained on MSO over $100$ randomly initialized episodes.
\end{itemize}

\subsection{Simulation Results}

Results in Fig.~\ref{fig:results} show that \textbf{RESPRECT} and \textbf{Residual} constantly outperform the success rate obtained with \textbf{Pre-Trained}. This demonstrates the effectiveness of the proposed RRL approach. 

\textbf{RESPRECT} manages to achieve in one million environment timesteps a comparable success rate as \textbf{G-PAYN} when the latter is trained for five million timesteps (we refer the reader to the results in App.~\ref{appendix:gpayn} for a comparison with the full training of \textbf{G-PAYN}), and outperforms it when the training is stopped after one million timesteps. We also show that our approach manages to achieve a comparable success rate as the \textbf{Demonstrations} baseline in twelve experiments out of fourteen, outperforming it in seven task instances.

Compared to fine-tuning and MetaRL based approaches for fast task adaptation of a pre-trained policy, overall, \textbf{RESPRECT} performs much better than \textbf{Fine-Tuning} and \textbf{Reptile}. The only exception is represented by the \textit{035\_power\_drill+VGN} experiment, where \textbf{Fine-Tuning} outperforms all the other considered methods, and the full training of \textbf{G-PAYN}. In those cases in which the baselines achieve a similar success rate to \textbf{RESPRECT}, they require a larger number of timesteps,  see for instance the \textit{010\_potted\_meat\_can+VGN} and the \textit{004\_sugar\_box+VGN} experiments.

In most of the experiments, for example the \textit{004\_sugar\_box+VGN}, the success rate of \textbf{RESPRECT} in the initial training timesteps has a steeper slope than \textbf{Residual}, which is crucial to speed-up the training procedure. Moreover, we noticed that \textbf{Residual} tends to have higher \textit{Critics} losses, which may lead to training instability. This occurs, for example in the \textit{021\_bleach\_cleanser+Superquadrics} experiment, where there is a performance drop after $\sim 300k$ timesteps.

\begin{figure*}
    \vspace{1mm}
    \centering
    \includegraphics[width=1\linewidth]{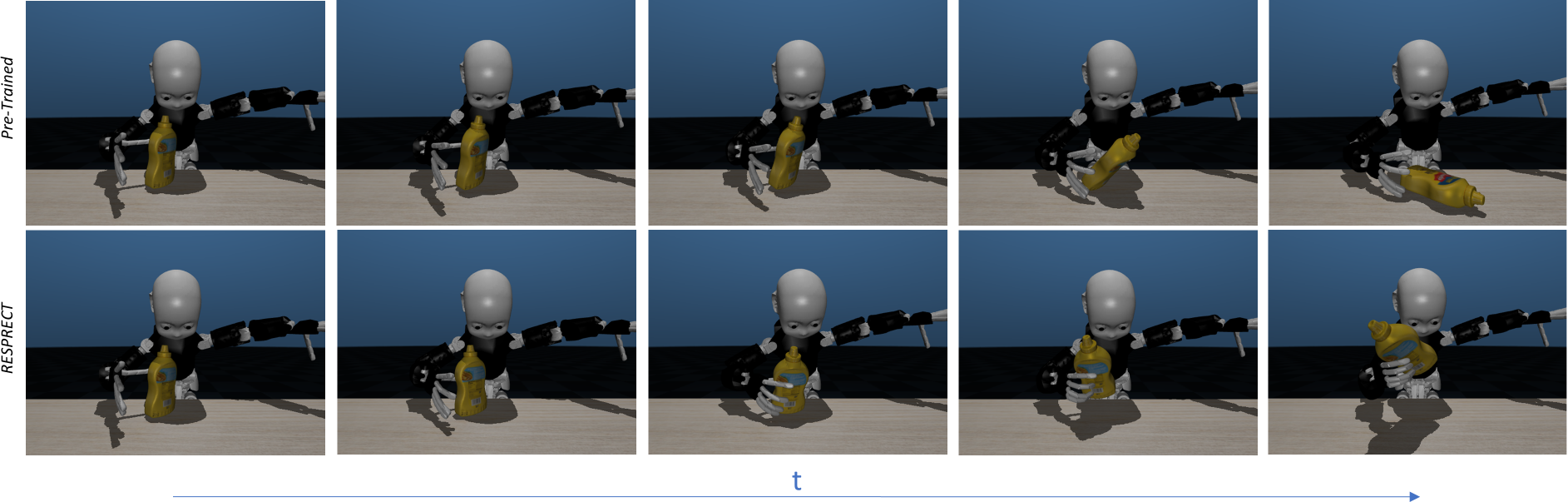}
    \caption{Qualitative evaluation of the proposed RESPRECT. We compare it to the pre-trained policy in the same experiment. We show how the residual output of RESPRECT allows to solve the task.}
    \label{fig:sim_pretr_respr}
\end{figure*}

In Fig.~\ref{fig:sim_pretr_respr}, we show a qualitative evaluation of RESPRECT in the iCub simulated environment. We report an exemplar sequence in which the residual policy successfully grasps the target object, while the pre-trained policy fails starting from the same object configuration. 

\subsection{Real Robot Results}

\begin{figure}
    \centering
    \includegraphics[width=1\linewidth]{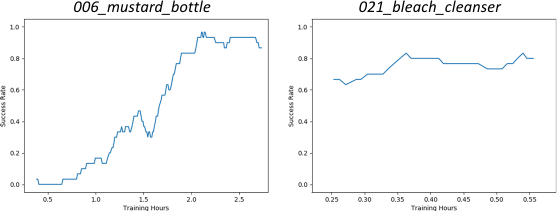}
    \caption{RESPRECT success rate (averaged over the last $30$ training episodes) for increasing training time on the real iCub robot.}
    \label{fig:icub_real}
\end{figure}

We train RESPRECT to grasp the \textit{006\_mustard\_bottle} and the \textit{021\_bleach\_cleanser} starting from grasp poses given by \textit{Superquadrics} on the real iCub humanoid. For these experiments, we consider as base policy the same policy pre-trained on the simulated MSO dataset used for the experiments in simulation. In Fig.~\ref{fig:icub_real}, we report the success rate for increasing robot training time. Results show that after $\sim2.5$ hours ($\sim10k$ timesteps) and $\sim30$ minutes ($\sim1.5k$ timesteps), the robot manages to successfully (with success rate $\sim 0.9$ and $\sim 0.8$) grasp the \textit{006\_mustard\_bottle} and the \textit{021\_bleach\_cleanser}. In Fig.~\ref{fig:icub_respr}, we show a successful grasping sequence during the \textit{006\_mustard\_bottle} training. We refer the reader to the video submitted as supplementary material\footnote{\label{video_fn}\url{https://youtu.be/JRsBLVclhpg}} which shows two successful grasps on the two considered objects and illustrates the whole training process for the \textit{006\_mustard\_bottle}.

%% file: sections/limitations.tex
With the proposed method, we managed to speed-up the training for a multi-fingered grasping task by a factor of $5$, while also removing the need for task demonstrations. However, training a model for grasping a single object for $1M$ timesteps still requires a considerable amount of time, especially if these must be executed on the real robot. 

From a qualitative analysis of the residual policies, we noticed that they struggle to react to failures during grasp execution. We believe that this is the reason why their overall success rate is comparable to the open-loop pipeline \textbf{Demonstrations}. We plan to further improve the reward function, adding components for adapting the position of the fingers once grasp failure is detected. Moreover, in the current problem setting, the observation of the environment comprises an estimate of the initial position of the object, which is kept constant throughout the grasping episode. We plan to integrate a class-agnostic tracker to keep updating the estimated position of the object. This can help obtaining more reactive grasping policies.

Finally, while we show that RESPRECT does not need a hand-tuned controller, and that this has some advantages with respect to the state-of-the-art, it is fair to say that our method requires a suitable base policy pre-trained with an \textit{Actor-Critic} DRL algorithm, such as SAC~\cite{haarnoja2018soft} or G-PAYN~\cite{ceola2023gpayn}.

\begin{figure*}
    \vspace{1mm}
    \centering
    \includegraphics[width=1\linewidth]{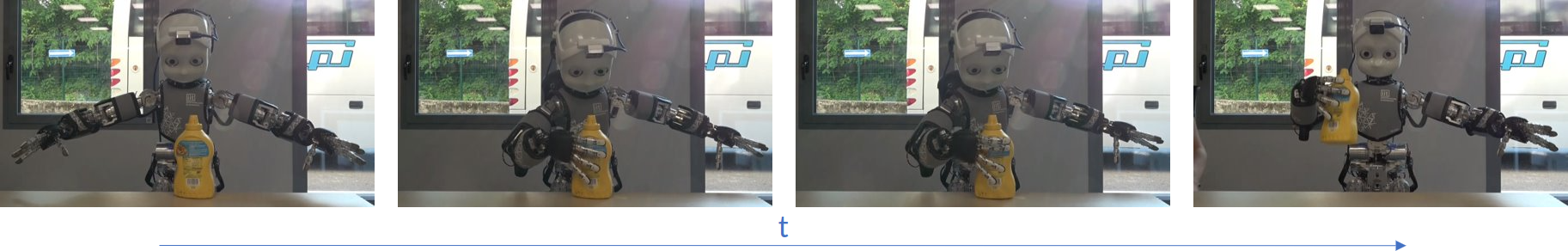}
    \caption{Object grasping with the iCub humanoid robot. In this exemplar sequence the policy was trained to grasp the \textit{006\_mustard\_bottle} with RESPRECT on the real robot.}
    \label{fig:icub_respr}
\end{figure*}

%% file: sections/conclusions.tex
Grasping with multi-fingered robotic hands is an important task for dexterous manipulation. State-of-the art DRL approaches that tackle this problem suffer from data-inefficiency and are difficult to deploy on real robots. Some recent works propose to train a policy only with simulated data, and to transfer the trained policy on the real robot without any adaptation. However, these approaches are highly dependent on the quality of the simulated environment, and may not be suitable for those cases in which there is a large sim-to-real gap. In addition, these approaches are intrinsically off-line and do not allow the robot to adapt after its deployment. In this perspective, we propose RESPRECT, with the aim of speeding-up the training of DRL policies to grasp novel objects. The proposed approach learns a residual policy for the object at hand on top of a policy pre-trained on a different set of objects. In contrast to existing RRL methods that leverage model-based controllers, we employ a pre-trained policy. This allows to use the weights of the latter to warm start some components of the residual policy to significantly speed-up the training.

We show that RESPRECT achieves a comparable success rate as G-PAYN~\cite{ceola2023gpayn} in a fraction of the training timesteps and without using task demonstrations. Moreover, RESPRECT outperforms two fine-tuning and MetaRL baselines for adaptation of a pre-trained policy on a new target object both in terms of success rate and training steps required to achieve comparable performance.

Finally, we deploy the proposed RESPRECT on the real iCub~\cite{icub} humanoid, showing that it is possible to obtain a policy that is trained directly on the real robot.

As a future work, we plan to improve the grasping pipeline to obtain policies which are more reactive to failures.

%% file: sections/appendix_ft.tex
\begin{figure}[!h]
    \vspace{-3.5mm}
    \centering
    \includegraphics[width=\linewidth]{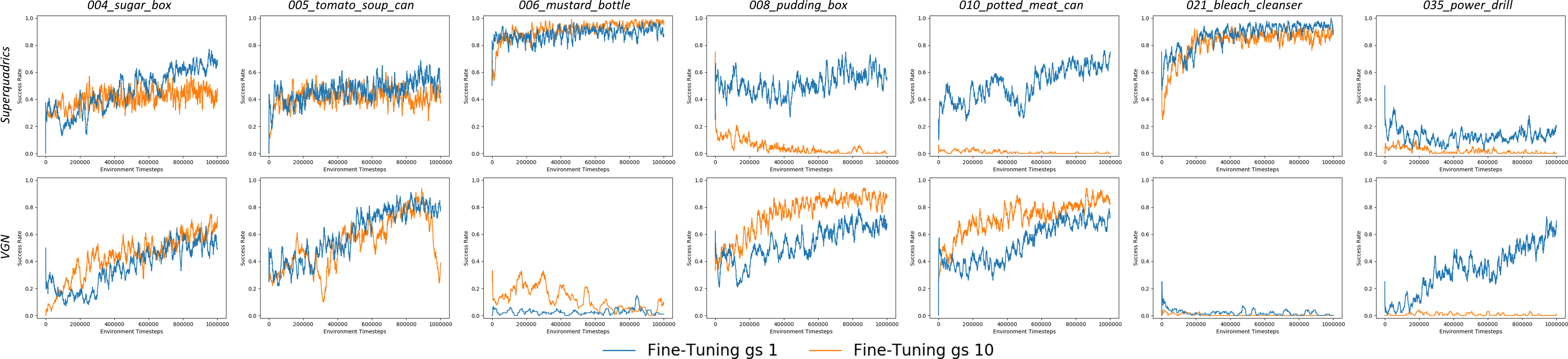}
    \caption{We evaluate \textbf{Fine-Tuning} considering one and ten gradient steps at each training timestep.}
    \label{fig:ft}
\end{figure}

In Fig.~\ref{fig:ft}, we compare the success rate achieved by \textbf{Fine-Tuning} updating the policies for one and ten gradient steps at each training iteration.

%% file: sections/appendix_reptile.tex
\begin{figure}[h]
    \vspace{-3.5mm}

    \centering
    \includegraphics[width=\linewidth]{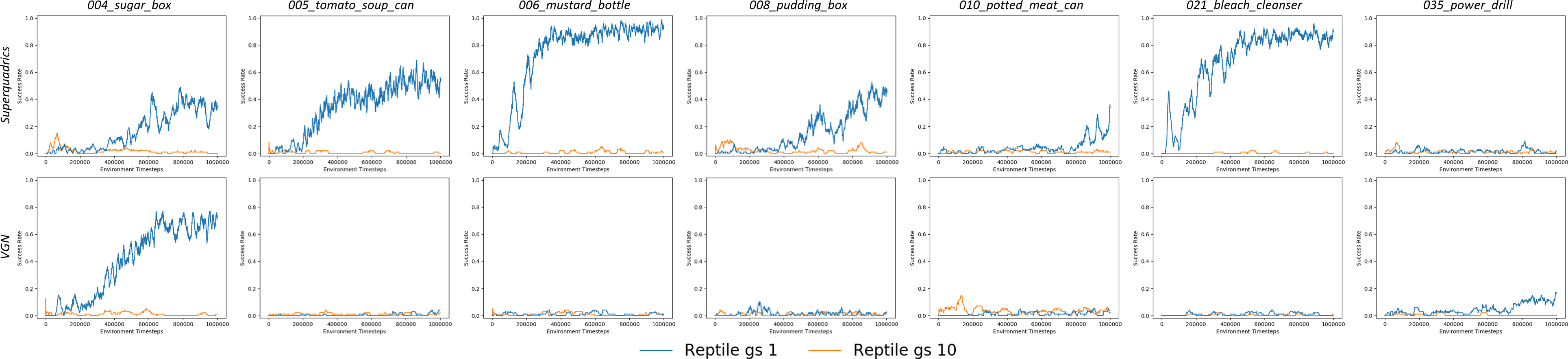}
    \caption{We evaluate \textbf{Reptile} considering one and ten gradient steps at each training timestep.}
    \label{fig:reptile}
\end{figure}

In Fig.~\ref{fig:reptile}, we evaluate \textbf{Reptile} updating the policies for one and ten gradient steps at each training iteration.

%% file: sections/appendix_gpayn.tex
\begin{figure}[h]
    \vspace{-3.5mm}

    \centering
    \includegraphics[width=1\linewidth]{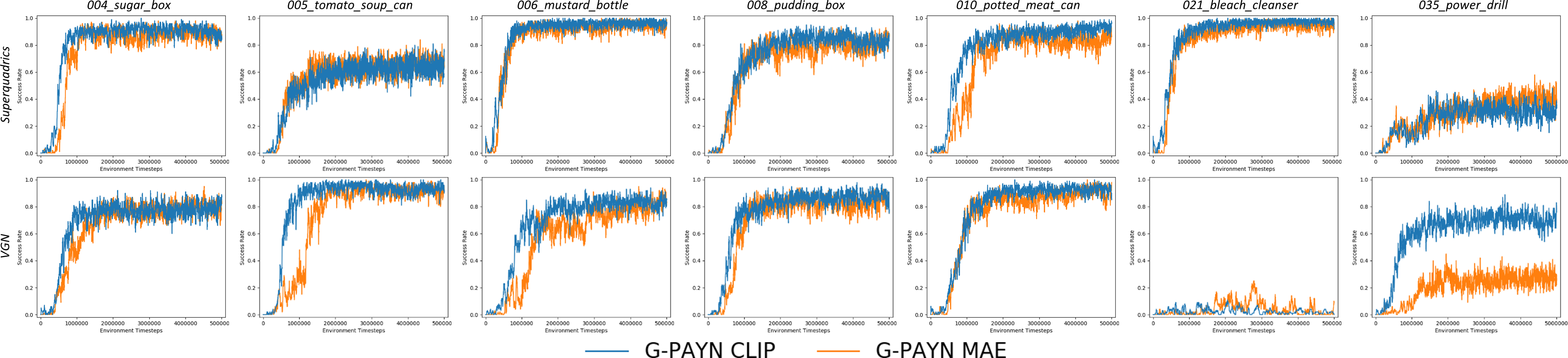}
    \caption{We evaluate \textbf{G-PAYN} trained with different visual backbones (MAE and CLIP) for $5M$ environment timesteps.}
    \label{fig:gpayn}
\end{figure}

In Fig.~\ref{fig:gpayn}, we compare the success rate achieved by \textbf{G-PAYN} trained with different visual backbones for $5M$ environment timesteps.

%% file: sections/appendix_residual.tex
\begin{figure}[h]
    \centering
    \includegraphics[width=\linewidth]{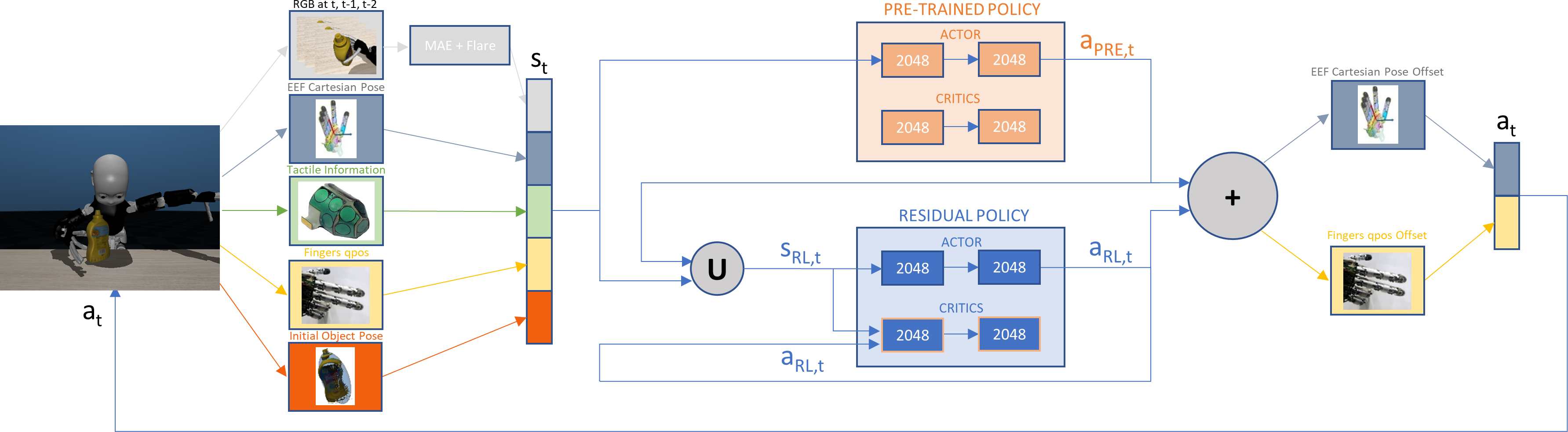}
    \caption{\textbf{Residual} overview. We compute state $s_t$ from RGB images at timesteps $t$, $t-1$ and $t-2$, end-effector cartesian pose, tactile information and finger joint poses. We compute action $a_t$ (composed of cartesian offsets for the end-effector and finger joint offsets) combining the outputs $a_{PRE, t}$ of the pre-trained policy and $a_{RL, t}$ of the residual policy. Note that $a_{RL, t}$ is the output of the residual policy \textit{Actor}, given the concatenation of $s_t$ and $a_{PRE, t}$ into $s_{RL, t}$. We train only the two $2048$-dimensional fully connected layers in the residual \textit{Actor} and \textit{Critics}. For the sake of clarity, we do not report the input of the \textit{Critics} in the pre-trained policy, and the output of both the \textit{Critics} being the same as the ones in SAC~\cite{haarnoja2018soft}.}
    \label{fig:pipeline_residual}
\end{figure}

In Fig.~\ref{fig:pipeline_residual}, we overview the \textbf{Residual} approach used to compare results obtained with \textbf{RESPRECT} in Sec.~\ref{sec:results}.

